\documentclass[10pt,twocolumn,letterpaper]{article}

\usepackage[pagenumbers]{cvpr} %
\usepackage{graphicx}
\usepackage{amsmath}
\usepackage{amssymb}
\usepackage{booktabs}
\usepackage{pifont}
\usepackage{multirow}
\newcommand{\cmark}{\ding{51}}%
\newcommand{\xmark}{\ding{55}}%
\usepackage[dvipsnames]{xcolor}
\usepackage[dvipsnames]{colortbl}
\usepackage{appendix}

\definecolor{Gray}{gray}{0.9}
\definecolor{LightCyan}{rgb}{0.88,1,1}

\newcolumntype{a}{>{\columncolor{Gray}}c}

\usepackage[pagebackref,breaklinks,colorlinks]{hyperref}

\usepackage[capitalize]{cleveref}
\crefname{section}{Sec.}{Secs.}
\Crefname{section}{Section}{Sections}
\Crefname{table}{Table}{Tables}
\crefname{table}{Tab.}{Tabs.}

\begin{document}


\title{$\text{Re}^\mathbf{2}$TAL: \underline{Re}wiring Pretrained Video Backbones for  \underline{Re}versible \\ \underline{T}emporal \underline{A}ction \underline{L}ocalization}

{\author{
Chen Zhao$^{1}$ 
\quad Shuming Liu$^{1}$
\quad Karttikeya Mangalam$^{2}$
\quad Bernard Ghanem$^{1}$
\and
$^{1}$King Abdullah University of Science and Technology (KAUST) \quad $^{2}$UC Berkeley\\
{\tt\small \{chen.zhao, shuming.liu, bernard.ghanem\}@kaust.edu.sa}
\quad {\tt\small mangalam@berkeley.edu}
}}
\maketitle

\begin{abstract}

Temporal action localization (TAL) requires long-form reasoning to predict actions of various durations and complex content. Given limited GPU memory, training TAL  end to end (i.e., from videos to predictions) on long videos  is a significant challenge. Most methods can only train on pre-extracted features without optimizing them for the localization problem, consequently limiting localization performance.
In this work, to extend the potential in TAL networks,
we propose a novel end-to-end  method \textbf{Re$^2$TAL}, which \underline{re}wires pretrained video backbones for \underline{re}versible  \underline{TAL}. \textbf{Re$^2$TAL} builds a backbone with reversible modules, where the input can be recovered from the output such that the bulky intermediate activations can be cleared from memory during training. 
Instead of designing one single type of reversible module, we propose a network rewiring mechanism, to \textbf{transform any module with a residual connection to a reversible module} without changing any parameters. 
This provides two benefits: (1) a large variety of reversible networks are easily obtained from existing and even future model designs, and
(2) the reversible models require  much less training effort as they reuse the pre-trained parameters of their original non-reversible versions.
Re$^2$TAL, only using the RGB modality, reaches 37.01\% average mAP on ActivityNet-v1.3, a new state-of-the-art record,  and mAP 64.9\% at tIoU=0.5   on THUMOS-14, outperforming all other RGB-only methods. Code will be available at \href{https://github.com/coolbay/Re2TAL}{https://github.com/coolbay/Re2TAL}.

\end{abstract}
\section{Introduction} \label{sec:into}

Temporal Action Localization (TAL)~\cite{ zhao2021video, lin2019bmn, ramazanova2022owl} is a fundamental problem of practical importance in video understanding. It aims to  
bound semantic actions within start and end timestamps. Localizing such video segments is very useful for 
a variety of tasks such as video-language grounding\cite{soldan2022mad, Hendricks2017LocalizingMI}, moment retrieval~\cite{grauman2022ego4d, chen2022internvideo}, video captioning~\cite{Krishna2017DenseCaptioningEI, Mun2019StreamlinedDV}. 
Since video actions have a large variety of temporal durations and  content, to produce high-fidelity localization, TAL approaches need to learn from a long  temporal scope of the video, which contains a large number of frames. To accommodate all these frames along with their network activations in GPU memory is extremely challenging,  given the current GPU memory size (\eg the commodity GPU GTX1080Ti only has 11GB). Often, it is impossible to train one video sequence on a GPU without substantially downgrading the video spatial/temporal resolutions.

\begin{figure}[t]
\begin{center}
\footnotesize
\includegraphics[width=0.45\textwidth]{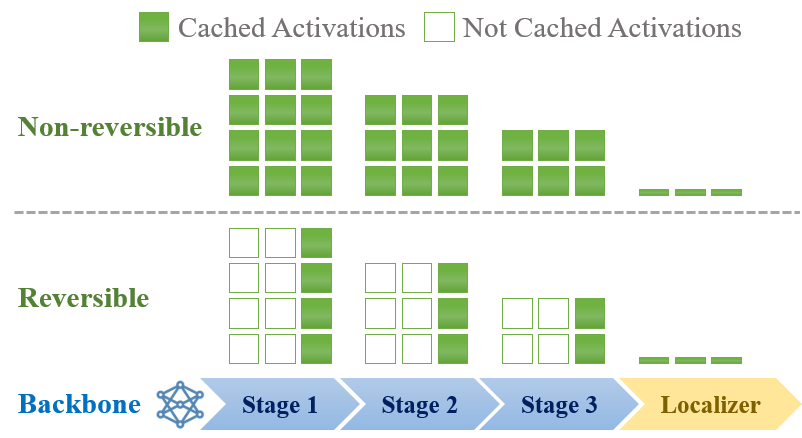}
\end{center}
\vspace{-15pt}
\caption{\textbf{Illustration of TAL network activations in training. }Top: non-reversible network stores activations of all layers in memory. Bottom: reversible network only needs to store the activations of inter-stage downsampling layers. Backbone activations dominate memory occupation, compared to Localizer.}
\vspace{-18pt}
\label{fig:gpu_mem}
\end{figure}

 To circumvent the GPU-memory bottleneck, most TAL methods deal with videos in two isolated steps (\eg \cite{bai2020boundary,  xu2021boundary, xu2020g,  zhao2021video, zhang2022actionformer, zhao2023segtad}). First is a snippet-level
feature extraction step, which simply extracts snippet representations using a  pre-trained video network (backbone) in inference mode. 
The backbone is usually a  large neural network trained  for  an auxiliary task on a large dataset of trimmed video clips (\eg, action recognition on Kinetics-400~\cite{Kay2017TheKH}). The second step trains a localizer on the pre-extracted features. In this way,  only the activations of the TAL head need to be stored in memory, which is tiny compared to those of the backbone (see the illustration of the activation contrast between backbone and localizer in Fig.~\ref{fig:gpu_mem}).
However, this two-step strategy comes at a steep price. 
The pre-extracted features can suffer from domain shift from the auxiliary pre-training task/data to TAL, and  do not necessarily align with the representation needs of TAL. This is because they cannot be finetuned and must be used as-is in their misaligned state  for TAL.  
A better alternative is to jointly train the backbone and localizer  end to end. But as mentioned earlier, the enormous memory footprint of video activations in the backbone makes it extremely challenging. Is there a way for end-to-end training without compromising data dimensionality?

Reversible networks~\cite{revnet, irevnet, revvit,revgraphnn} provide an elegant solution to drastically reduce the feature activation memory during training. Their input can be recovered from the output via a reverse computation. Therefore, the intermediate activation maps, which are used for back propagation, do not need to be cached during the forward pass (as illustrated in Fig.~\ref{fig:gpu_mem}).
This offers a promising approach to enable memory-efficient end-to-end TAL training, and various reversible architectures have been proposed, such as RevNet~\cite{revnet}, and RevViT~\cite{revvit}. 
However, these works design a specific reversible architecture and train for a particular dataset. Due to their new architecture, they also need to train the networks from scratch, requiring
a significant amount of compute resources.

Conversely, it would be beneficial to be able to convert existing non-reversible video backbones to reversible ones, which would \textbf{(1)} avail a large variety of architectures and \textbf{(2)}  allow us to reuse the large compute resources that had already been invested in training the non-reversible video backbones. 
Since pre-trained video backbones are a crucial part of TAL, the ability to convert off-the-shelf backbones to reversible ones is a key to unleash their power in this task.

In this work, for end-to-end \underline{TAL}, we propose a principled approach to \underline{Re}wire the architectural connections of a pre-trained non-reversible backbone to make it \underline{Re}versible, dubbed Re$^2$TAL. Network modules with a residual connection (res-module for short), such as a Resnet block~\cite{he2016deep} or a Transformer MLP/attention layer~\cite{ViT}, are the most popular design recently. \textbf{Given any network composed of  residual modules, we can apply our rewiring technique to convert it to a corresponding reversible network} without introducing or removing any trainable parameters. Instead of training from scratch, our reversible network can reuse the non-reversible network's parameters and only needs a small number of epochs for finetuning to reach similar performance.   
We summarize our contributions as follows.

(1) We propose a novel approach to construct and train reversible video backbones parsimoniously by architectural rewiring from an off-the-shelf pre-trained  video backbone. This not only provides a large collection of reversible candidates, but also allows reusing the large compute resources invested in pre-training these models. We apply our rewiring technique to various kinds of representative video backbones, including transformer-based Video Swin and ConvNet-based Slowfast, and  demonstrate that our reversible networks can reach the same performance of their non-reversible counterparts with only minimum finetuning effort (as low as 10 epochs compared to 300 epochs for training from scratch).

(2) We propose a novel approach for  end-to-end TAL training using reversible video networks. Without sacrificing spatial/temporal resolutions or network capability, our proposed approach dramatically reduces GPU memory usage, thus enabling end-to-end training on one 11GB GPU. We demonstrate on different localizers and different backbone architectures that we significantly boost TAL performance with our end-to-end training compared to traditional feature-based approaches.

(3) With our proposed Re$^2$TAL, we use recent localizers in the literature to achieve a new state-of-the-art performance,  $37.01\%$ average mAP on ActivityNet-v1.3. We also reach the highest mAP among all methods that only use the RGB modality on THUMOS-14,  $64.9\%$ at tIoU$=0.5$, outperforming concurrent work TALLFormer \cite{cheng2022tallformer}.

\section{Related Work} \label{sec:related_work}

\noindent{\textbf{Reversible Networks}} are a family of neural models based on the reversible real-valued non-volume preserving (real NVP) transformation introduced in \cite{NICE, dinh2017density}. The transformation has been originally applied widely for image generation using generative flows~\cite{flow++, glow}, and later used for various applications, such as compression~\cite{liu2021semantics},  denoising~\cite{liu2021invertible}, and steganography recovery~\cite{Mou2023LFVSN}. Furthermore, it has also been repurposed for memory-efficient neural network training for a variety of architectures such as ConvNets~\cite{revnet, irevnet}, Masked ConvNet~\cite{rev_maskconv}, RNNs~\cite{revrnn},  Graph Networks~\cite{revgraphnn} and more recently, Vision Transformers~\cite{revvit}.
However, each of these methods only focuses on one or several specific architectures, and trains their newly proposed reversible architectures from scratch, thereby requiring large compute resources. In this work, we propose a rewiring scheme to adapt off-the-shelf models to reversible architectures which only utilize a small amount of compute for finetuning. This allows democratizing the reach of reversible architecture to arbitrary models and dataset settings by leveraging the architecture design effort and computational cost already spent in vanilla off-the-shelf models.  

\vspace{4pt}\noindent{\textbf{Video Recognition Backbones.}} ConvNets have had a long and illustrious history of improving video understanding performance~\cite{Simonyan2014,Feichtenhofer2016, Qiu2017, Li2018, Xie2018,Tran2019, Wu2019, girdhar2019video, feichtenhofer2020x3d, Zhou2017, jiang2019stm, slowfast, carreira2017quo}. Among those, Slowfast~\cite{slowfast}, which uses a slow pathway and a fast pathway to capture spatial semantics and temporal motion respectively, has been widely adopted for various  tasks such as action detection~\cite{slowfast}, untrimmed video classification~\cite{liu2022fineaction} and moment/natural language retrieval~\cite{grauman2022ego4d} for its high efficiency and efficacy. 
Recently, video transformers have ushered in a new life in the field of video representation learning. Leveraging long-term temporal connection via the attention mechanism,  transformers have quickly gain popularity as the backbone of choice for several video recognition workloads~\cite{vivit, mvit, mvitv2, mtv, timesformer, vswin}. Among those, Video Swin Transformer~\cite{vswin} introduces an inductive bias of locality to video transformers, leading to an outstanding speed-accuracy trade-off. However, most of the models are vanilla non-reversible architectures. In this work, we unleash the potential of reversible architectures for these models such that the good properties of reversible models, \eg memory efficiency, can be infused into them. 
In particular, we show that the pretrained Video Swin Transformer~\cite{vswin} and  SlowFast~\cite{slowfast} models can be rewired to be reversible, and finetuned cheaply to improve temporal action localization performance with end-to-end training. 

\vspace{4pt}\noindent{\textbf{Temporal Action Localization (TAL).}} 
Due to the conflict between large  video data and the GPU memory limit, most TAL methods using deep networks are two-step methods~\cite{Shou2016TemporalAL,lin2018bsn,lin2019bmn,qing2021temporal,zhao2021video, chao2018rethinking, Long2019GaussianTA, lin2019fast, liu2020tsi, bai2020boundary,zhao2023segtad,xu2020g,zhang2022actionformer}. For example, using pre-extracted features, G-TAD~\cite{xu2020g} and VSGN~\cite{zhao2021video} utilize graph convolutions to model temporal relations between snippets, and ActionFormer~\cite{zhang2022actionformer}  leverages transformers to capture long-range context. To mitigate the performance gap between the two-step mechanism and real end-to-end training, some methods explore post-pretraining to enhance the video feature representations for TAL~\cite{alwassel2021tsp,xu2017r,xu2021low}.
In the meanwhile, researchers also attempt to perform real end-to-end  training by reducing network/data complexity~\cite{xu2017r,liu2022empirical, liu2022etad,wang2021rgb,liu2020progressive,lin2021learning,cheng2022tallformer}. R-C3D~\cite{xu2017r} is the end-to-end pioneer, but it uses a shallow network C3D~\cite{tran2015learning}, thus restricting the performance. PBRNet~\cite{liu2020progressive} and ASFD~\cite{lin2021learning} downscale the frame resolution to $96$$\times$$96$. DaoTAD~\cite{wang2021rgb} makes RGB-only enough by using end-to-end training with various data augmentations. 
TALLFormer\cite{cheng2022tallformer} proposes to use a feature bank strategy and only updates a portion of features during end-to-end training. 
Compared to these approaches, our method doesn't sacrifice any data dimensionality or data samples in training, and it supports very deep networks. We significantly reduce memory consumption while preserving the full data fidelity. Moreover, our work is complementary to these methods, and can also be used jointly to further reduce memory cost.

\section{Method: Re$^2$ Temporal Action Localization} \label{sec:method}

\subsection{TAL Formulation and Architecture} \label{sec:tal_formulation}
Temporal action localization (TAL) predicts the timestamps of actions from a video sequence, which is formulated as follows. Given a video sequence $V$ of $T$ frames $\{I_t$ $\in$$ \mathbb{R}^{3\times H \times W}\}_{t=1}^T$, TAL predicts a set $m$ of actions $\Phi=  \left \{ \phi _m=\left ({t}_{m, s},{t}_{m, e}, {c}_m, {s}_m \right ) \right \}_{m=1}^{M}$, where ${t}_{m, s}$ and ${t}_{m, e}$ are  action start and end time respectively, ${c}_m$ is action label, and ${s}_m$ is prediction confidence. To achieve this, the following two steps are required. 

In the first step, the videos $V$ are  encoded as $N$ features vectors $\{\mathbf{x}_n $$\in$$ \mathbb{R}^{C}\}_{n=1}^N$ via  a backbone. This backbone
aggregates spatial information within video frames,  as well as temporal information across frames. It is usually designed for an auxiliary task such as action recognition. Popular backbones can be categorized into Resnet-based architectures, such as I3D~\cite{carreira2017quo}, R2+1D~\cite{tran2018closer}, SlowFast~\cite{slowfast}, and Transformer based architectures, such as ViViT~\cite{vivit}, Video Swin Transformers~\cite{vswin}. 

In the second step, a localizer, uses a  `neck' to further  aggregate the video features $\{\mathbf{x}_n\}_{n=1}^N$ in the temporal domain, and a `head', to make 
 predictions of action boundaries, \ie, start and end timestamps $({t}_{m, s},{t}_{m, e})$ and categories. The neck contains layers of networks, \textit{e.g.}, 1D convolutional networks in BMN~\cite{lin2019bmn}, Graph networks in G-TAD~\cite{xu2020g} and VSGN~\cite{zhao2021video}, and Transformers in RTD-Net~\cite{tan2021relaxed} and ActionFormer~\cite{zhang2022actionformer}.

As mentioned in Sec.~\ref{sec:into}, an optimal way to train the entire TAL network is to jointly train the backbone and the localizer end to end. However, it is substantially challenging to fit all  activation maps of the long video sequence into limited GPU memory. In the following sections, we provide  end-to-end TAL solution Re$^2$TAL.

\subsection{Rewire for Reversibility, and Reuse }\label{sec:rewire}
Let's first analyze what in the GPU memory precludes end-to-end training of TAL. As mentioned in Sec.~\ref{sec:into}, the activation maps in the backbone are the major occupant in the memory. Concretely, given a batch of a video sequence of $T$ frames, each with resolution $S$$\times$$S$, to train them on a backbone of $L$ layers each with $C$ channels, the training memory complexity  is $\mathcal{O}(LCTS^2)$. Downgrading any data dimensionality
or  the backbone capacity can reduce the memory complexity, but at the same time, may harm the prediction accuracy. We aim to reduce  memory consumption without sacrificing the data dimensionality or the model capacity. 

Actually, the intermediate activations are stored in memory during training  for the purpose of  gradient computation in back propagation. If we can reconstruct them during back propagation, then there is no need for their memory occupation. Reversible networks~\cite{revnet, irevnet,  revvit, Mou2023LFVSN} is a superb solution for reconstructing the input from output, and various reversible architectures have been proposed recently (e.g., Revnet~\cite{revnet}, RevViT~\cite{revvit}). However, the performance of TAL heavily depends on the backbone that is pre-trained on one or even multiple large-scale datasets. For example, the popular I3D~\cite{carreira2017quo} backbone is trained on Kinetics-400~\cite{Kay2017TheKH} with its parameters initialized from Resnet~\cite{he2016deep} trained on ImageNet~\cite{deng2009imagenet}. If we make up a new reversible network for TAL as is the case with previous reversible networks (e.g., \cite{revnet, irevnet}), we need to  go through all the expensive and time-consuming training stages to reach an equivalently good initialization for TAL.

Instead of proposing a new reversible architecture, we propose a principled approach to rewire a pre-trained non-reversible backbone 
to make it reversible. Given   \textit{any}  module with a residual connection, we can convert it into a reversible module while preserving the same set of  parameters. This technique has two advantages. First, we obtain a large and growing collection of various reversible architectures out of the box, by rewiring the existing networks and even future networks of higher performance. 
Second, we can reuse their pre-trained weights to initialize our reversible networks, eliminating the tremendous effort to train from scratch. In the following, we will describe our rewiring for reversibility (\textbf{Re}$^{\textbf{2}}$) technique in detail.

\begin{figure}[t]
\begin{center}
\footnotesize
\includegraphics[width=0.45\textwidth]{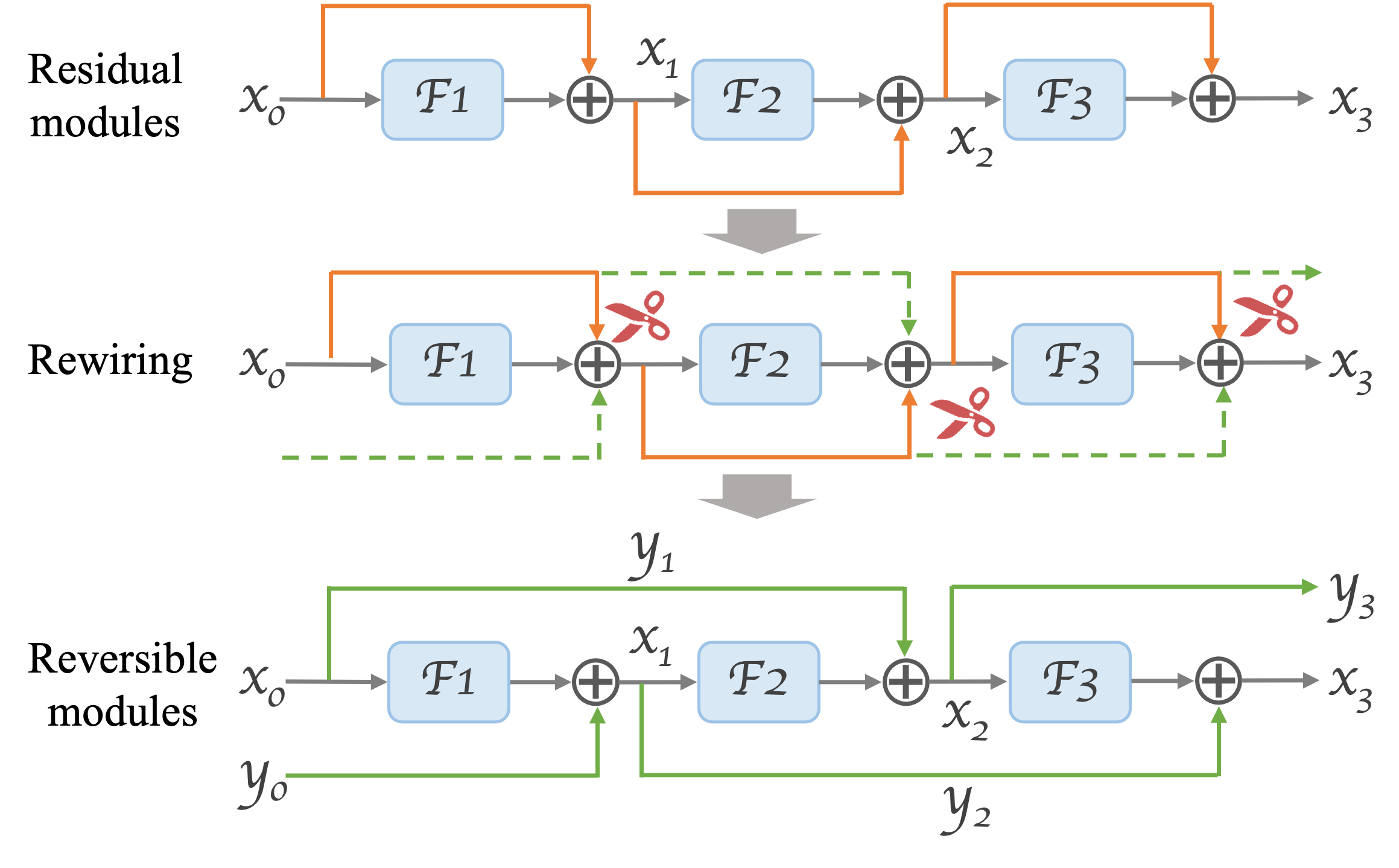}
\end{center}
\vspace{-20pt}
\caption{\textbf{Illustration of the proposed rewiring process.} In this example,  3 consecutive residual modules in the same stage  are rewired into 3 reversible modules. After rewiring, each residual connection skips one more $\mathcal{F}$ block ahead. Note that beyond this example, our rewiring can process any number of such modules.}
\vspace{-18pt}
\label{fig:rewire}
\end{figure}

\begin{figure}[t]
\begin{center}
\footnotesize
\includegraphics[width=0.45\textwidth]{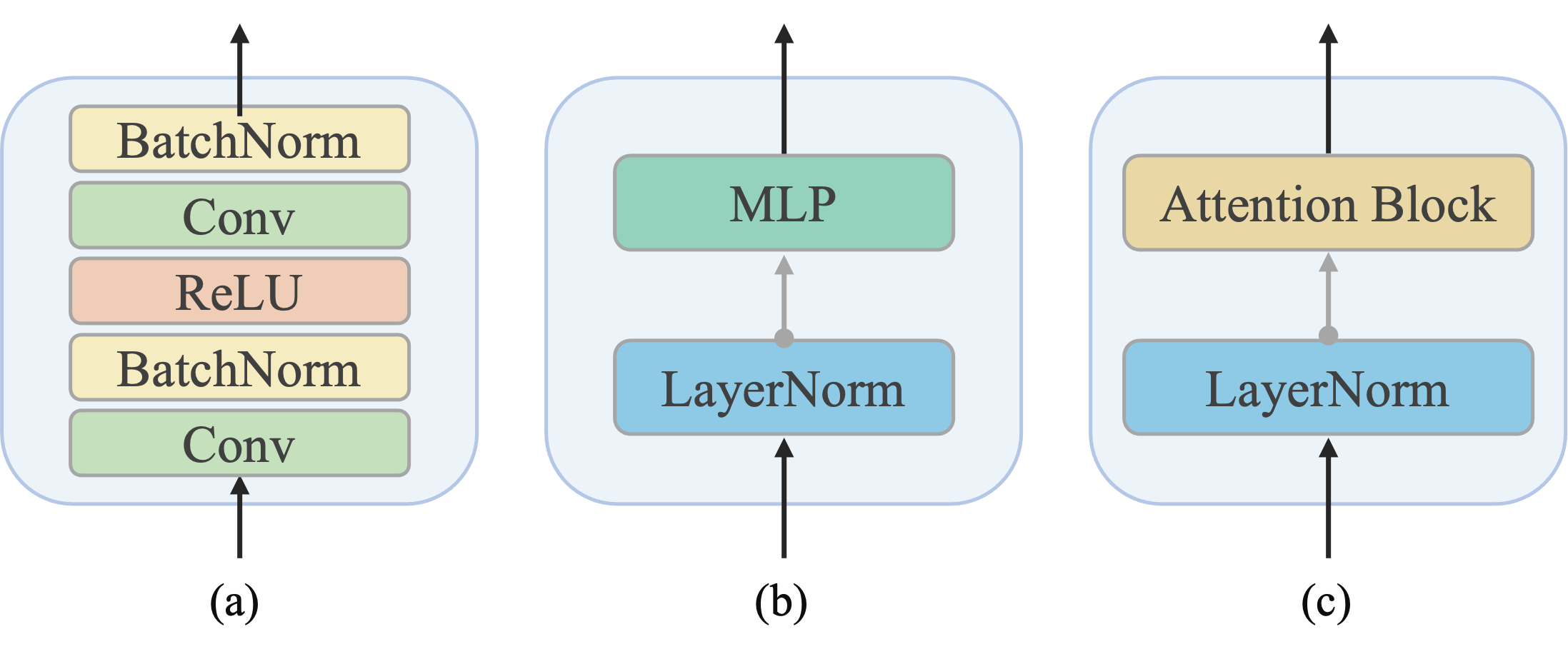}
\end{center}
\vspace{-15pt}
\caption{\textbf{Examples of $\mathcal{F}$ blocks.} (a) A resnet basic block. (b) A transformer MLP layer. (c) A transformer attention layer.}
\vspace{-15pt}
\label{fig:f_block}
\end{figure}

\textbf{Rewiring}. Residual modules, \textit{i.e.}, modules with a residual connection, are the most commonly adopted design in concurrent neural network architectures, such as Resnet~\cite{he2016deep} and Transformers~\cite{ViT}. Given any residual modules, we can rewire them into reversible modules. Fig.~\ref{fig:rewire} illustrates this process with an example of 3 consecutive residual modules that are in the same stage, \ie, there are no downsampling operations in between. Each of the original residual modules contains a block $\mathcal{F}_i$ (blue boxes) where $i=1,2,3, \ldots$  and a residual connection (orange arrow).
The blocks $\mathcal{F}_i$ have the same  input and output dimensions, and can contain any computations.
For example, in Resnet, the building block consisting of convolution, batch normalization, and ReLU can be one $\mathcal{F}_i$ block (\eg Fig.~\ref{fig:f_block} (a)). In Transformers, if $\mathcal{F}_i$ represents an attention layer, then  $\mathcal{F}_{i+1}$ is an MLP layer (\eg Fig.~\ref{fig:f_block} (b) and (c)). 
The residual connection in each module only skips one $\mathcal{F}_i$ block in the same module.

To rewire, we simply let the residual connection skip the next block $\mathcal{F}_{i+1}$  as well as the current one $\mathcal{F}_{i}$. To be more specific, we keep the starting point of each residual connection, but make it end after two blocks, as shown in the second row of Fig.~\ref{fig:rewire}. As a result, two consecutive residual connections become overlapped, and there are always two pathways of activations (gray and green in Fig.~\ref{fig:rewire}) throughout the modules, as shown in the third row of Fig.~\ref{fig:rewire}. 

To prepare a second pathway of input to the first module, we duplicate the input $x_0$ to obtain $y_0$. To combine the results $x_3$ and $y_3$ of the two pathways, we simply average them at the end of the modules. This design guarantees that the dimensions of the input and output of every block $\mathcal{F}_i$ are identical before and after the rewiring. 
Therefore, the structure of $\mathcal{F}_i$ stays exactly the same.

\begin{figure*}[t]
\begin{center}
\footnotesize
\includegraphics[width=0.8\textwidth]{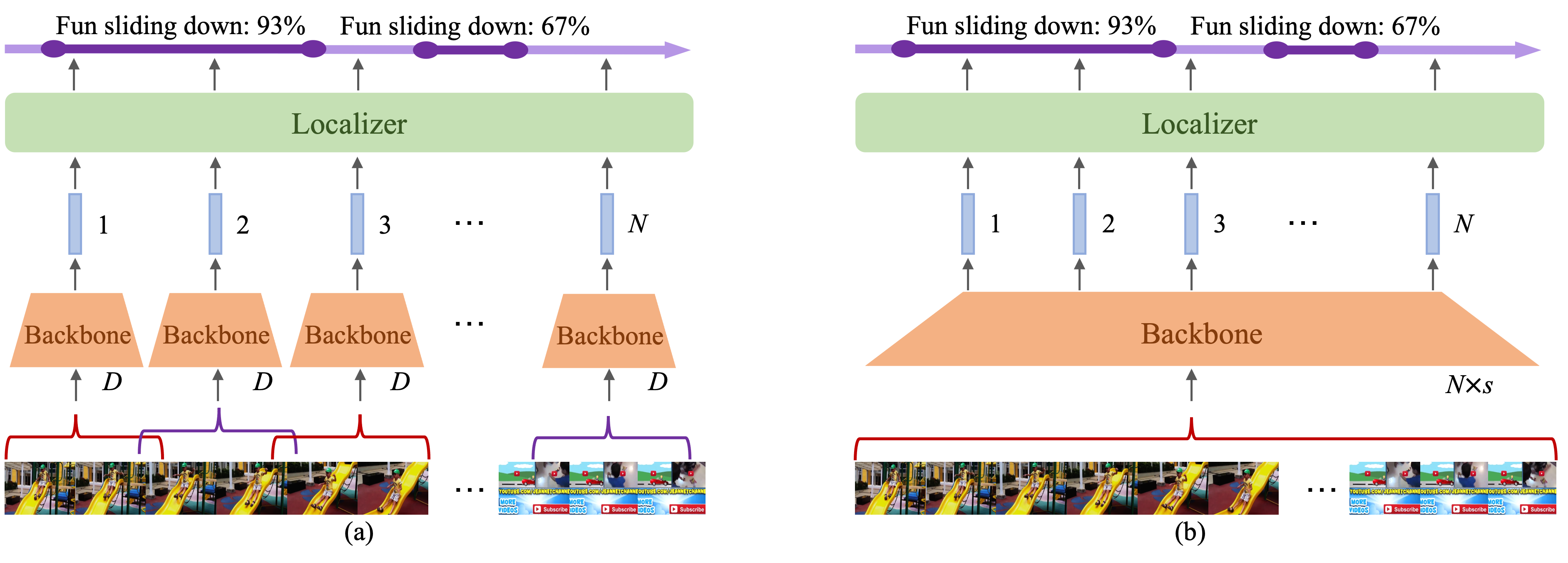}
\end{center}
\vspace{-20pt}
\caption{\textbf{Comparison of different input arrangements for a TAL network.}  \textbf{(a) Snippets input:} the backbone processes many video snippets (a short clip of frames), often overlapped. \textbf{(b) Frames input:} the backbone takes all the frames in the video as one single input.}
\vspace{-17pt}
\label{fig:snippet_frame}
\end{figure*}

\textbf{Reversibility}. In the following, we mathematically formulate the rewired modules to show their reversibility. To be concise, we use the first two modules as an example. Given the input activations $x_0$ and $y_0$, and the  blocks $\mathcal{F}_1$ and $\mathcal{F}_2$ , the output of the two blocks are $x_1, y_1$ and $x_2, y_2$ respectively, computed as follows
\begin{equation}
\left\{\begin{array} { l }
{ y _ { 1 } = x _ { 0 }  } \\
{ x _ { 1 } = \mathcal{F }_1 ( x_{ 0 } ) + y_{ 0 } },
\end{array} \Rightarrow \left\{\begin{array}{l}
y_2=x_1 \\
x_2=\mathcal{F}_2(x_1)+y_1.
\end{array}\right.\right.
\label{eq:forward}
\end{equation}
The above equations are  reversible, which means that we can recover the input $x_1, y_1$ from $x_2, y_2$, and then $x_0, y_0$ from $x_1, y_1$. 
The reverse computation is formulated in the following equation 
\begin{equation}
\left\{\begin{array}{l}
x_0=y_1 \\
y_0=x_1-\mathcal{F}_1(x_0),
\end{array}
\Leftarrow
\left\{\begin{array} { l }
{ x _ { 1 } = y _ { 2 } } \\
{ y _ { 1 } = x _ { 2 } - \mathcal { F }_2 ( x_ { 1 } ) }.
\end{array}\right.\right.
\label{eq:reverse}
\end{equation} 

A network with more modules follows the same strategy as in Eq.~\ref{eq:forward} and Eq.~\ref{eq:reverse}. 
When we stack multiple consecutive reversible modules as one network (as the example in Fig.~\ref{fig:rewire}), the entire network is reversible. In this case, we can start from the last module and sequentially reconstruct the input of each module with Eq.~\ref{eq:reverse}.

\textbf{Reuse}. As the structure and parameter dimensions of the $\mathcal{F}_i$ block in our reversible modules after rewiring stay exactly the same as the $\mathcal{F}_i$ block in the corresponding residual module, we can directly reuse the pre-trained parameters in the residual modules to initialize our reversible modules.

Our rewiring mechanism   provides a large collection of reversible candidates. We can easily convert a network into a reversible one without designing a new architecture from scratch. Not only can we use the existing architectures, but we can also benefit from future even better models to obtain better reversible models.
The reuse strategy  allows taking advantage of the large computing resources invested in those pre-trained models. We don't need to re-train the reversible network from scratch. 
Instead, we can  train the reversible backbone with minimum effort (as low as 10 epochs compared to 300 epochs from scratch) while reaching the same performance as its residual counterpart.

\subsection{Reversible Temporal Action Localization  }

Considering that the backbone is the heaviest part in the TAL network in terms of memory usage, as illustrated in Fig.~\ref{fig:gpu_mem}, we target the backbone and apply our Re$^2$ technique described above to rewire  it to obtain a reversible network. 

A backbone network, \eg, Video Swin Transformer~\cite{vswin}, Slowfast~\cite{slowfast},  is usually comprised of several stages. Within each stage, the activation sizes stay the same. We convert all the modules into reversible ones following the rewiring method proposed in Sec.~\ref{sec:rewire}. During training, we can clear all the input and intermediate activations from GPU memory, and only store the final output of the stage during the forward pass (as shown in Fig.~\ref{fig:gpu_mem}). In the back propagation, we re-compute all the activations based on the reverse process as in Eq.~\ref{eq:reverse}. As \cite{revvit} pointed out, this re-computation doesn't incur too much more computational time since we can parallelize the process to make use of the spare computation of GPUs. 

Across stages, there are downsampling layers, which reduce the activation sizes. We leave the downsampling layers as they are and cache the activations inside to enable back propagation. Considering that there are only several downsampling layers, the memory occupation of these activations is acceptable.

\subsection{End-to-End TAL Training}
For end-to-end training with our Re$^2$TAL, we just need to do the following: find a well-performing video backbone, rewire it into a reversible one, load the parameters from the original backbone to the reversible one and finetune for several epochs on the pretraining task, and train with a localizer end to end. 
This reversible TAL network is significantly more efficient in memory usage, enabling end-to-end  training a  GPU of limited memory (\eg, as small as a 11GB commodity GPU). But is it a wise choice to directly adopt the training strategies from the feature-based methods?

\textbf{Input frame arrangement}. 
Since most TAL methods are designed and experimented with the pre-extracted, they have predisposed to particular design choices, such as extracting features with overlapped snippets to arrange the input frames. But we find it not ideal for end-to-end training. 

Fig.~\ref{fig:snippet_frame}~(a) illustrates the framework commonly used in  two-step (or feature-based) methods. To extract $N$ feature vectors as the localizer input, $N$ snippets  need to be processed by the backbone independently. Each snippet contains a sequence of $D$ frames. 
The snippets are usually overlapped with one another to maintain temporal consistency, which causes duplicate computation and extra memory occupation. Consequently, at least $ND$ frames need to go through the backbone.  

This framework works fine with the two-step method, since feature extraction is one-off effort and all the snippets can be processed sequentially to circumvent the memory issue. However,  end-to-end training cannot bear the extra memory cost. Therefore,  we treat the entire video sequence as one single input instead, and the backbone aggregates all frames at one time, as shown in Fig.~\ref{fig:snippet_frame}~(b). This mechanism not only reduces the  cost incurred by the duplicate computation, but also has the advantage of aggregating long-term temporal information, in contrast to being restricted within the snippet as with the snippet mechanism. In this way, to obtain $N$ feature vectors, we just need to process $Ns$ frames, where $s$ is the overall strides in the backbone (\eg $s$$=$$2$ in Video Swin Transformers). Considering that $s$ is usually much smaller than $D$, the frame-input mechanism is more efficient in memory and computation cost.

\section{Experiments} \label{sec:exp}

\textbf{Datasets and evaluation metrics.} We present our experimental results on two representative datasets {ActivityNet-v1.3} (ActivityNet for short)~\cite{caba2015activitynet} and {THUMOS-14} (THUMOS for short)~\cite{jiang2014thumos}.  \textbf{ActivityNet} is a challenging large-scale dataset, with $19994$ temporally annotated untrimmed videos in $200$ action categories, which are split into training, validation and testing sets by the ratio of $2$:$1$:$1$.  \textbf{THUMOS} contains $413$ temporally annotated untrimmed videos with $20$ action categories, in which $200$ videos are for training and $213$ videos for validation\footnote{The training and validation sets of THUMOS are temporally annotated videos from the validation and testing sets of UCF101~\cite{UCF101}, respectively.}.   For both datasets, we use mean Average Precision (mAP) at different tIoU thresholds as the evaluation metric. On ActivityNet,  we choose $10$ values in the range $[0.5, 0.95]$ with a step size $0.05$ as tIoU thresholds; on THUMOS, we use tIoU thresholds $\{0.3, 0.4, 0.5, 0.6, 0.7\}$;  following the official evaluation practice.

\textbf{Backbones and localizers of Re$^2$TAL.} For backbones, we choose two representative models from the Resnet and Transformer families respectively for experimental demonstration: Video Swin Transformers (Vswin for short)~\cite{vswin} and Slowfast~\cite{slowfast}, both well known for their powerful video representation and memory efficiency. For Vswin, we use three variants: tiny, small, and base; for Slowfast, we use  50, 101, and 152.
For localizers, we experiment  with the recent temporal action localization (TAL) methods VSGN~\cite{zhao2021video} and ActionFormer~\cite{zhang2022actionformer}.

\textbf{Implementation Details of Re$^2$TAL.} We initialize all our reversible backbones with their non-reversbile counterparts, and finetune for up to $30$ epochs on Kinetics-400 with Cosine Annealing learning rate policy and Adamw (for Vswin) and SGD (for Slowfast) optimizers. Actually, in our experiments, we find $10$ epochs of finetuning already reaches similarly good performance. With the pre-trained reversible backbone, we train  TAL  on a \textbf{single} GPU, A100 with batchsize $=1$ for Vswin and V100 with batchsize $=2$ for Slowfast. This is only possible due to the massive GPU memory savings from the reversible backbones.  For the hyper-parameters, we follow the original training recipe of VSGN~\cite{zhao2021video} and ActionFormer~\cite{zhang2022actionformer} for the learning rates and epochs in the localizers. We set the learning rates of the Vswin backbones  $1$ magnitude lower than the  localizer, and those of the Slowfast backbones $2$ magnitude lower. The spatial resolution is $224\times224$. For ActivityNet, the number of frame input is $T=512$, the number of feature vectors is $N=256$;   and for THUMOS, the number of frame input is $T=1024$, the number of features is $N=512$.

\begin{table}[t]
\centering
\caption{\textbf{Advantage of  end-to-end training.} End-to-end training leads to significant mAP (\%) boost for TAL with either backbone.} 
\vspace{-5pt}
\setlength{\tabcolsep}{2.0pt}
\small
\begin{tabular}{l|ccca|ccca} 
\toprule
 &\multicolumn{4}{c|}{Re$^2$Vswin-tiny} &\multicolumn{4}{c}{Re$^2$Slowfast-101}\\
\hline
Method &  0.5 & 0.75  & 0.95  & Avg.& 0.5 & 0.75  & 0.95  & Avg. \\
\hline
Features &{51.18}&{35.09}&{9.72}& 34.47 &51.98&36.00&9.47&35.24\\  
\textbf{End2End} &\textbf{53.24}&\textbf{37.23}&\textbf{10.49}&\textbf{36.38} &\textbf{53.63}&\textbf{37.53}&\textbf{10.67}& \textbf{36.82}\\  
\bottomrule
\end{tabular}
\vspace{-15pt}
\label{tab:e2e_benefit_anet}
\end{table}

\subsection{Ablation and Analysis}
In this section, we provide ablation study on ActivityNet and performance analysis to answer the following questions. (1) Why do we need end-to-end training? (2) How effective and efficient is our Re$^2$TAL? (3) What is the benefit of the proposed rewiring technique? (4) Which reversible backbone and localizer provides the best performance?

\subsubsection{Why end-to-end training?} 

With end-to-end training, we are able to optimize the  features to adapt to the TAL task and dataset such that we can achieve higher performance than the feature-based method. 
To verify this, we compare end-to-end training  
to the feature-based training on our reversible TAL. 
In Tab.~\ref{tab:e2e_benefit_anet}, we demonstrate the comparison  on two types of backbones: Re$^2$ Vswin-tiny and Re$^2$ Slowfast-101 with VSGN~\cite{zhao2021video} on ActivityNet.
It shows that for both backbones, using   end-to-end training leads to  significant performance boost, gaining almost $2\%$ mAP.

\begin{figure}[t]
\begin{center}
\footnotesize
\includegraphics[width=0.45\textwidth]{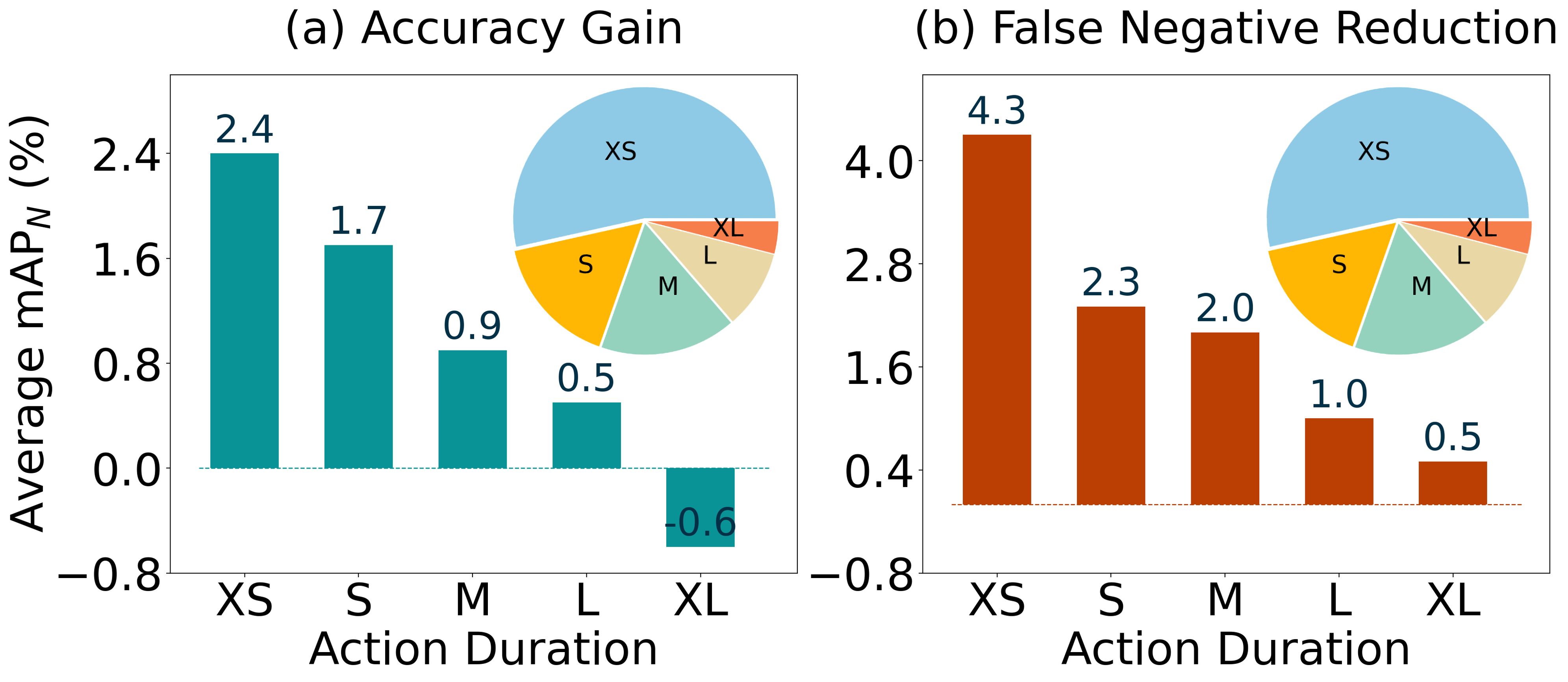}
\end{center}
\vspace{-15pt}
\caption{\textbf{Performance improvement of end-to-end TAL upon feature-based in terms of action temporal duration}. }
\vspace{-12pt}
\label{fig:detad}
\end{figure}
To further diagnose what kinds of actions end-to-end training improves the most, in Fig.~\ref{fig:detad} (a), we plot the accuracy gains brought by end-to-end training for five different groups of actions based on their temporal durations (in seconds): XS: (0s, 30s],  S: (30s, 60s], 
M: (60s, 120s],  L: (120s, 180s], and XL: $>$ 180s (as suggested in \cite{detad}). We can see that  the accuracy gains for different action durations are closely related to the numbers of samples in each duration category (the pie chart): short actions (XS  and X) with many samples are obviously improved, though long actions (XL) with fewer samples is slightly sacrificed.   
Actually, short actions are a fundamental challenge in TAL~\cite{zhao2021video}. End-to-end training provides an effective solution by enabling the backbone to learn feature representations to the benefit of short actions --- the categories with more samples, thus preventing them fusing into the background. Short actions occupy the majority in the dataset, and their improvement leads to an overall performance boost.
In Fig.~\ref{fig:detad} (b), we show that the false negative predictions are reduced by end-to-end training for all groups, more significantly for shorter ones.

\subsubsection{How effective and efficient is Re$^2$TAL?}

\begin{figure}[t]
\begin{center}
\footnotesize
\includegraphics[width=0.48\textwidth]{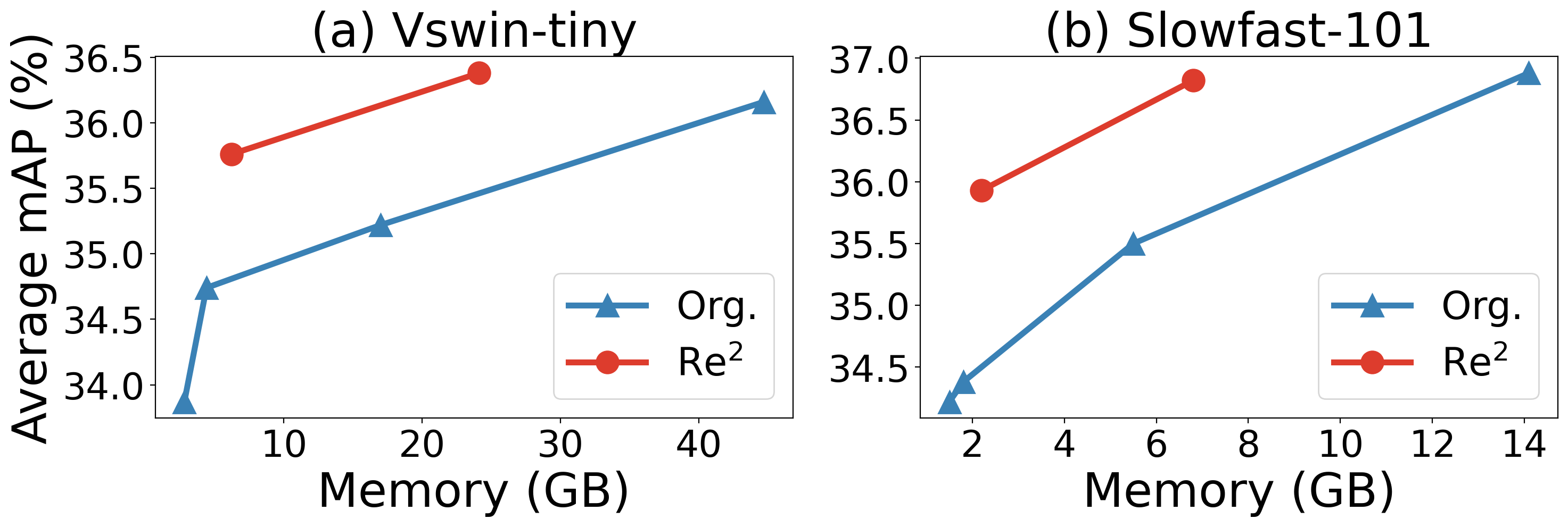}
\end{center}
\vspace{-15pt}
\caption{\textbf{ Re$^2$TAL models compared with original models at different spatial and temporal downscaling ratios.} Spatial and temporal resolutions from left to right are as follows. Re$^2$TAL: (112, 512), (224, 512); Original: (56, 512), (112, 192), (224, 192), (224, 512). The localizer VSGN~\cite{zhao2021video} is used.}
\vspace{-10pt}
\label{fig:cmp_downsampling}
\end{figure}

An alternative way to enable end-to-end training is to reduce data resolutions~\cite{liu2020progressive, wang2021rgb}. In Fig.~\ref{fig:cmp_downsampling}, we downscale the input videos in the spatial or temporal dimensions to adjust the memory requirements, and compare our Re$^2$TAL models to their corresponding original models. 
 We can see  that under even smaller memory requirements, our Re$^2$TAL models achieve higher performance  than the original models that 
 relies on sacrificing video resolutions to reduce memory.

\begin{figure}[t]
\begin{center}
\footnotesize
\includegraphics[width=0.48\textwidth]{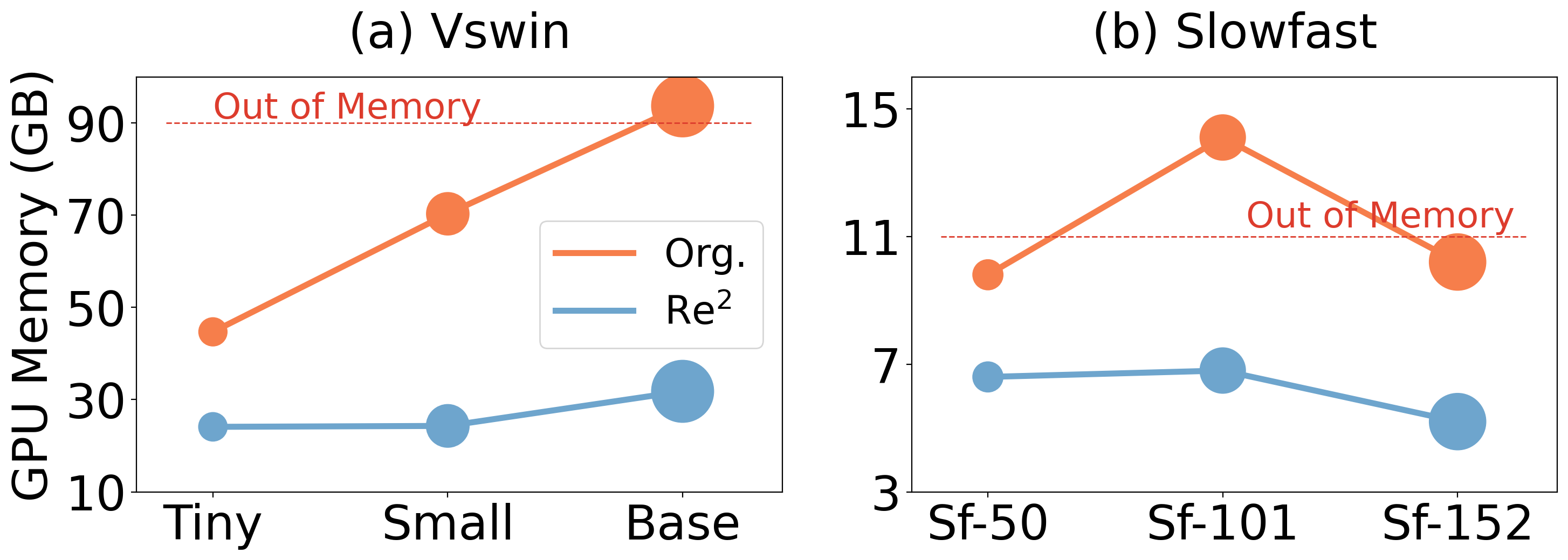}
\end{center}
\vspace{-15pt}
\caption{\textbf{GPU memory consumption v.s. network sizes. } The dot sizes represent the model sizes, and the y-axis is the GPU memory used for processing one video (batchsize$=$$1$). Our Re$^2$TAL almost keeps the memory constant when only the network depths increases (\eg from Vswin-tiny to small, from Slowfast-50 to 101), whereas the original models easily go out of memory when the network becomes larger. }
\vspace{-10pt}
\label{fig:memcomparison}
\end{figure}

To further visualize the memory efficiency of our Re$^2$TAL compared to their non-reversible counterparts when trained end to end, we demonstrate the GPU memory consumption of the three Vswin backbones of different depths and widths and the three Slowfast backbones of different depths in Fig.~\ref{fig:memcomparison}. For Vswin backbones, we assume a GPU memory budget of 90GB, which is the size of A100. For Slowfast backbones, we assume our GPU budget is 11GB, which is the size of the commodity GPU such as GTX1080Ti. We can see that for either case, the non-reversible network will go out of memory when the model size reaches a certain level. In contrast, our Re$^2$TAL almost keeps constant memory usage  when only the network depth increases (from Vswin-tiny to small, from Slowfast-50 to 101)\footnote{Memory increases from Vswin-small to base is due to the channel increase; memory reduction from Slowfast-101 to 152 is because the former one uses input configuration (8, 8) while the latter uses (4, 16). }.
With the low memory cost,  our Re$^2$TAL enables end-to-end training with a deep Slowfast backbone on \textbf{one single} 11GB GPU. Moreover, Re$^2$TAL doesn't incur much extra training time compared to its non-reversible counterpart. The time to train one epoch is the following, Vswin-tiny ($114$ mins) \vs Re$^2$Vswin-tiny ($135$ mins); Slowfast-50 ($147$ mins) \vs Re$^2$Slowfast-50 ($158$ mins).

\subsubsection{Why rewiring and reuse?} 
We rewire existing network architectures to make them reversible, and reuse their parameters for initialization. This way we dramatically reduce the  effort of training the reversible networks, and still  reach the representation capability of the original non-reversible ones.  To compare the video representation capabilities of both types of networks, we use them to extract video features and train a localizer (VSGN~\cite{zhao2021video} in this case) with the features. We demonstrate their performance in Tab.~\ref{tab:cmp_rev_org}, 
showing that our reversible models are comparable to the original models.

\begin{table}[t]
\centering
\caption{\textbf{Comparison of feature representations  between the Re$^2$TAL and   original models}, in terms of average mAP (\%) on the dataset ActivityNet. Vw: Vswin; Slowf: Slowfast. }
\vspace{-7pt}
\setlength{\tabcolsep}{3pt}
\small
\begin{tabular}{l|ccc|l|ccc}
\toprule
Model &Tiny & Small & Base &Model& 50 & 101 & 152 \\
\hline
Vswin &34.36 &33.86 &\textbf{34.47} & Slowfast & 34.60 &35.04 &\textbf{34.61}\\
Re$^2$Vw  &\textbf{34.47} &\textbf{34.04} &34.09 & Re$^2$Slowf &\textbf{34.93} &\textbf{35.24} &34.54 \\

\bottomrule
\end{tabular}
\vspace{-8pt}
\label{tab:cmp_rev_org}
\end{table}

\begin{table}[t]
\centering
\caption{\textbf{Effectiveness of rewiring and reusing pre-trained video models}.  Reusing pre-trained models leads to significantly better performance than training from scratch (compare Row 1 to the rest). Reusing a better pre-trained model gives even higher performance (compare Row 2 to Row 3). Pret.: pretraining.}
\vspace{-3pt}
\small
\setlength{\tabcolsep}{4.8pt}
\begin{tabular}{ll|ccc|a}
\toprule
 Pret. Model & Pret. Dataset  & 0.5 & 0.75 & 0.95&  Avg. \\
\hline
Vswin-base &None &44.58&28.41&7.40& 28.68\\
 & Kinetics-400 &52.72 &36.73&8.88&35.73 \\
 & Kinetics-600 &\textbf{52.46}&\textbf{37.37}&\textbf{10.39}& \textbf{36.28} \\
\bottomrule
\end{tabular}
\vspace{-8pt}
\label{tab:rewire_reuse}
\end{table}

\begin{table}[t]
\centering
\caption{\textbf{Comparison of localization performance with different backbones and localizers}, in terms of average mAP (\%) on the dataset ActivityNet. 
}
\vspace{-5pt}
\small
\setlength{\tabcolsep}{3.0pt}
\begin{tabular}{ll|ccc|a}
\toprule
Backbone & Localizer &  0.5 & 0.75 & 0.95 & Avg. \\ 
\hline

Re$^2$Vswin-tiny  &VSGN &53.24&37.23&10.49 &36.38\\ 
  &ActionFormer &54.75&37.81&9.03 & \textbf{36.80} \\ 
\hline
Re$^2$Slowfast-101   &VSGN &{53.63} &{37.53}&{10.67}& {36.82}\\ 
  & ActionFormer&{55.25}&{37.86}&{9.05} & {\textbf{37.01}} \\
\bottomrule
\end{tabular}
\vspace{-10pt}
\label{Tab:bb_localizer}
\end{table}

\begin{table*}[t]
\centering
\caption{\textbf{Compared to the state-of-the-art for temporal action localization performance on ActivityNet-v1.3 and THUMOS-14}, measured by  mAPs (\%) at different tIoU thresholds according to their respective official metrics. E2E: end-to-end; Mem: memory (GB).
}
\vspace{-5pt}
\small
\setlength{\tabcolsep}{3.5pt}
\begin{tabular}{l|ccc|c|ccca|ccacc}
\toprule
\multirow{2}{*}{\textbf{Method}} & \multicolumn{1}{c}{\multirow{2}{*}{\textbf{Backbone}}} & \multicolumn{1}{c}{\multirow{2}{*}{\textbf{E2E}}} & \multirow{2}{*}{\textbf{Flow}} & \multirow{2}{*}{\textbf{Mem}} & \multicolumn{4}{c|}{\textbf{ActivityNet-v1.3 }}                                                       & \multicolumn{5}{c}{\textbf{THUMOS-14}}                                        \\
\cline{6-9} 
\cline{10-14}
                   &               &            &                      &                      & 0.5 & 0.75 & 0.95 & Avg. & 0.3 & 0.4 & 0.5 & 0.6 & 0.7 \\
\hline
TAL-Net~\cite{chao2018rethinking} & I3D &\xmark &\cmark&- &38.23 & 18.30 & 1.30 & 20.22 &53.2 & {48.5} & {42.8} & {33.8} & 20.8 \\ 
BMN~\cite{lin2019bmn}&TSN&\xmark&\cmark& - & {50.07} & {34.78} & {8.29} & {33.85} &{56.0} & 47.4 & 38.8 & 29.7 & 20.5  \\
{G-TAD}~\cite{xu2020g}  & TSN&\xmark&\cmark&- &{50.36} & {34.60} & {9.02} & {34.09} &{54.5} & {47.6} & {40.2} & {30.8} & {23.4}   \\
TSI~\cite{liu2020tsi} &TSN &\xmark&\cmark&- & 51.18 & 35.02 & 6.59 & 34.15 & 61.0 & 52.1 & 42.6 &  33.2 & 22.4 \\
BC-GNN~\cite{bai2020boundary} &TSN&\xmark&\cmark&-&50.56 & 34.75 & {9.37} & 34.26 & 57.1 & 49.1 & 40.4 & 31.2 & 23.1 \\
{VSGN~\cite{zhao2021video}} &TSN &\xmark&\cmark& - &{52.38}  & {36.01}  & 8.37 &{35.07} & {66.7} &  {60.4} & {52.4} & {41.0} & {30.4} \\
ActionFormer~\cite{zhang2022actionformer} &I3D&\xmark&\cmark&- &53.50 & 36.20 &8.20 & 35.60 &82.1 & 77.8 & 71.0 &59.4 &43.9 \\
PBRNet~\cite{liu2020progressive} &I3D&\cmark& \cmark&-&{53.96} & 34.97 & 8.98 & 35.01 &58.5 & 54.6 & {51.3} & {41.8}  & {29.5}  \\
AFSD~\cite{lin2021learning} &I3D&\cmark &\cmark&12 &52.40 & 35.30 & 6.50& 34.40 &67.3 &62.4 &55.5 & 43.7 & 31.1 \\
\hline
\hline
R-C3D~\cite{xu2017r}  &C3D&\cmark&\xmark& - &26.80 & - & - & -  &44.8 & 35.6 & 28.9 & - & -  \\
DaoTAD~\cite{wang2021rgb}  & I3D &\cmark & \xmark & 11 & - & - & - & - & 62.8 & - & 53.8 & - & 30.1 \\
TALLFormer~\cite{cheng2022tallformer}&VSwin-Base &\cmark &\xmark & 29&54.10 &36.20& 7.90& 35.60 &76.0 &-&\underline{63.2}&-& 34.5 \\
\hline
ActionFormer~\cite{zhang2022actionformer} & VSwin-Tiny&\xmark &\xmark& -&53.83 &35.82 &7.27  & 35.17 &70.8&64.7 &55.7  &42.2& 27.0 \\
\textbf{ActionFormer + Re$^2$TAL}&Re$^2$VSwin-Tiny&\cmark &\xmark &24 &{\underline{54.75}}& {\underline{37.81}}& {\underline{9.03}}& {\underline{36.80}} &{\underline{77.0}}&{\underline{71.5}  }&{ 62.4}& {\underline{49.7}} &{\underline{36.3}}\\ 
\hline
ActionFormer~\cite{zhang2022actionformer}  &Slowfast-101&\xmark&\xmark&-  &53.98 &37.00 &8.87 &36.09 &72.7  &66.9 &58.6  &46.4& 33.1 \\
\textbf{ActionFormer + Re$^2$TAL }&Re$^2$Slowfast-101&\cmark &\xmark & 6.8 &{\textbf{55.25}}&{\textbf{37.86}}&{ \textbf{9.05}} & {\textbf{37.01}} & {\textbf{77.4}} &{\textbf{72.6}} &{\textbf{64.9}}&{\textbf{53.7}}&{\textbf{39.0}} \\
\bottomrule
\end{tabular}
\vspace{-8pt}
\label{tab:sota}
\end{table*}

If there are two versions of the same non-reversible model trained in different ways and with different performance, will our Re$^2$TAL models benefit more from the higher-performing version? Vswin-base happens to have such two versions: one trained on Kinetics-400 and the other trained on Kinetics-600, the latter with better performance. We initalize our Re$^2$Vswin-base with either version, and compare their performance on TAL in Tab.~\ref{tab:rewire_reuse}.
By using the better version, the TAL performance obviously increases. In addition, using either version as initialization is better than training from scratch. This indicates that our rewiring-reuse strategy is very important  for TAL performance, and reusing a better pre-trained model will benefit TAL even further. That means if a better training strategy for an existing model comes out, we can correspondingly obtain a more accurate reversible model.

\subsubsection{Choice of Backbones and Localizers}
In Tab.~\ref{Tab:bb_localizer}, we demonstrate the results using  the two different backbones Re$^2$Vswin-tiny and  Re$^2$Slowfast-101, and two different localizers VSGN~\cite{zhao2021video} and ActionFormer~\cite{zhang2022actionformer} on ActivityNet. Since the ActionFormer localizer yields better performance for both Re$^2$Vswin-tiny and Re$^2$Slowfast-101, we compare the ActionFormer results to other methods in the literature, and also apply them  to the THUMOS dataset, as shown in Sec.~\ref{sec:SOTA}.

\subsection{State-of-the-Art Comparisons} \label{sec:SOTA}
We compare the performance of our Re$^2$TAL to recent state-of-the-art (SOTA) methods in the literature in Tab.~\ref{tab:sota} on ActivityNet and THUMOS.
On ActivityNet, our Re$^2$TAL reaches a new SOTA  performance: average mAP $37.01\%$, outperforming all other methods by significant margins. On THUMOS, ours surpasses the concurrent work TaLLFormer~\cite{cheng2022tallformer}  with mAP $64.9\%$ at tIoU$=$$0.5$, and the highest among all  methods that only use the RGB modality.

Furthermore, for an apple-to-apple comparison with the feature-based method ActionFormer, we re-ran ActionFormer using their official code with the RGB features extracted with the Vswin-tiny and Slowfast-101 backbones, corresponding to our Re$^2$Vswin-tiny and Re$^2$Slowfast-101, respectively. We see that our Re$^2$TAL always outperforms vanilla ActionFormer under the same backbone categories (\eg  Re$^2$Vswin-tiny and  Vswin-tiny  are in the same backbone category; Re$^2$Slowfast-101 and Slowfast-101 are in the same backbone category).

\section{Conclusions} \label{sec:conclusion}

In this work, we propose a novel rewiring-to-reversibility  (Re$^2$) scheme to convert off-the-shelf models into reversible models while preserving the number of trainable parameters. The procedure allows reusing the  compute invested in training large models and adds only a tiny sliver of finetuning compute. We apply the procedure to video backbones such as Video Swin (Vswin) and SlowFast to obtain Re$^2$Vwin and Re$^2$Slowfast backbones respectively. Further, we utilize the Re$^2$ backbones for memory-efficient end-to-end temporal action localization, reaching mAP $64.9\%$ at tIoU$=0.5$ on THUMOS-14, and average mAP $37.01\%$ on ActivityNet-v1.3, establishing a new state-of-the-art. We hope that future work in this direction can explore extending the Re$^2$ method to other memory-bottlenecked tasks such as dense video captioning, movie summarization \etc. 

\noindent\textbf{Acknowledgments}. This work was supported by the King Abdullah University of Science and Technology (KAUST) Office of Sponsored Research through the Visual Computing Center (VCC) funding and SDAIA-KAUST Center of Excellence in Data Science and Artificial Intelligence.
\newpage
\appendix
\section{Supplementary Material}

\maketitle

In this supplementary material, we discuss about data augmentation with our Re$^2$TAL (Section~\ref{sec:augmentation}) and our strategy for a special case of Resnet modules ( Section~\ref{sec:explain}).

\subsection{Data Augmentation}\label{sec:augmentation}

Data augmentation, especially spatial augmentation is a common strategy to  overcome overfitting. By randomly transforming the images when training a deep neural network, it is helpful to enhance the validation accuracy. However, it is rarely used for temporal action localization (TAL) due to the fact that most methods are based on 1D features,  which have already lost the spatial dimensions.

Due to the end-to-end training framework, we can take advantage of the various augmentations in the image domain as well as the temporal domain. In Table~\ref{tab:augmentation}, we compare the performance with different augmentation strategies: random spatial cropping, random flipping, temporal jittering, random rotation, and color jittering. We can see that the augmentations in the image domain are very important for improving performance. Using the combination of all the above augmentations leads to the highest performance.

\begin{table}[ht]
\centering
\caption{\textbf{Ablation study on different augmentations.} Performance is shown in mAP (\%) on Re$^2$ Vswin-tiny. } 
\setlength{\tabcolsep}{2.6pt}
\small
\begin{tabular}{ccccc|ccc|a} 
\toprule
Rand.  & Rand.  & Temp.  & Rand.  & Color &\multicolumn{3}{c|}{tIoU}& Avg. \\\cline{6-8}
Crop. &  Flip. &  Jitter. &  Rotat. &  Jitter. &0.5 &0.75&0.95&mAP\\
\hline
&&& &&52.27 &35.96&9.61&35.25 \\
\hline
\cmark&\cmark&& & & 52.59 &36.74&10.03&35.87 \\ 
\cmark&\cmark&\cmark& & & 53.05 &\underline{36.99}&9.08&35.98 \\ 
\cmark&\cmark&&\cmark& &\underline{53.16} &36.80&\textbf{10.55}& \underline{36.20}\\ 
\cmark&\cmark& &&\cmark &52.93 &{36.94}&10.17&36.02 \\ 
\cmark&\cmark&\cmark&\cmark&\cmark &\textbf{53.24} &\textbf{37.23}&\underline{ 10.49 }& \textbf{ 36.38} \\
\bottomrule
\end{tabular}
\label{tab:augmentation}
\end{table}

\subsection{Additional Explanations of A Special Case}\label{sec:explain}

\begin{figure}[ht]
\begin{center}
\footnotesize
\includegraphics[width=0.48\textwidth]{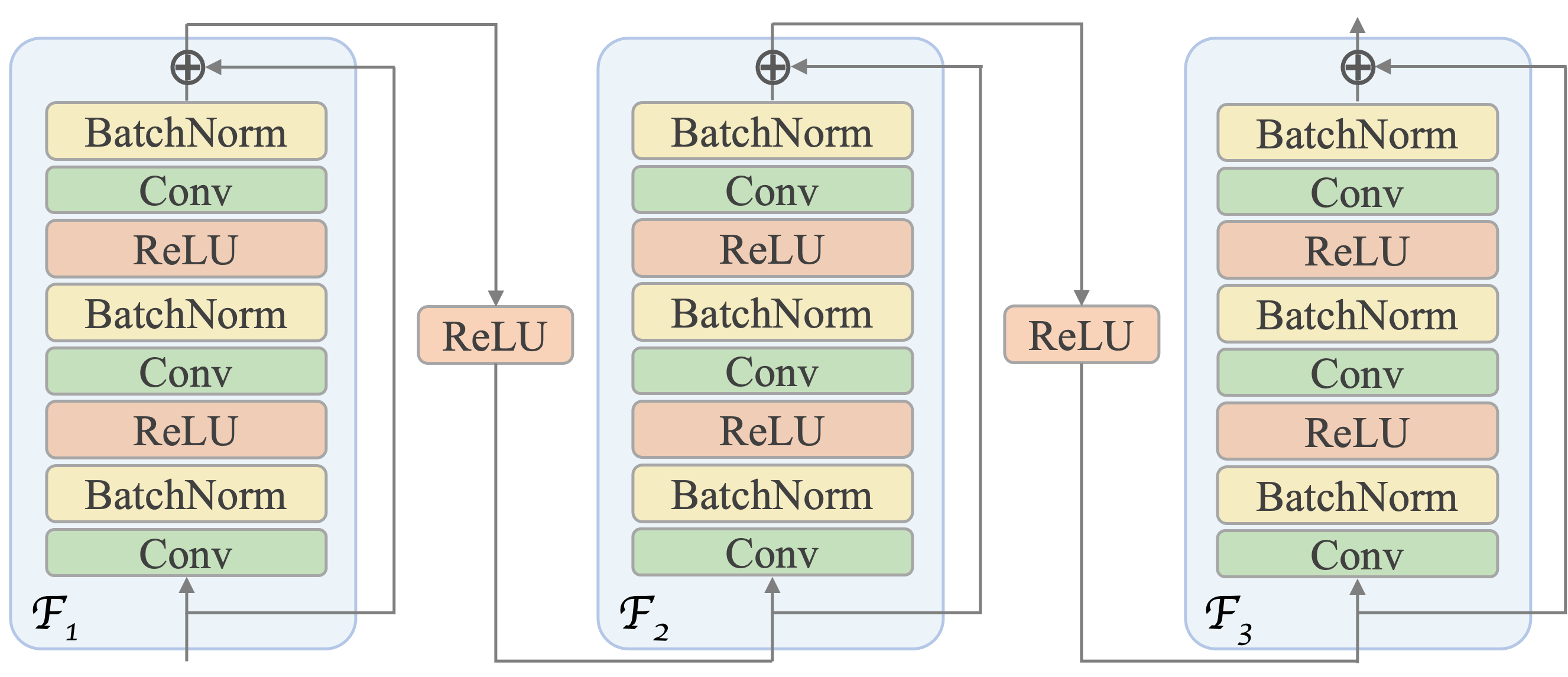}
\end{center}
\caption{\textbf{Illustration of $\mathcal{F}_i$ blocks (with bottleneck) with ReLU layers in between.}}

\label{fig:resnet_block}
\end{figure}

In the original version of the Resnet module~\cite{he2016deep}, there is a ReLU layer after each residual connection, between two $\mathcal{F}_i$ blocks, as illustrated in Fig.~\ref{fig:resnet_block}. Considering that the ReLU layer is not invertible,  directly rewiring this type of modules  would discontinue the reversibility across modules without caching the activations.

\begin{figure}[ht]
\begin{center}
\footnotesize
\includegraphics[width=0.48\textwidth]{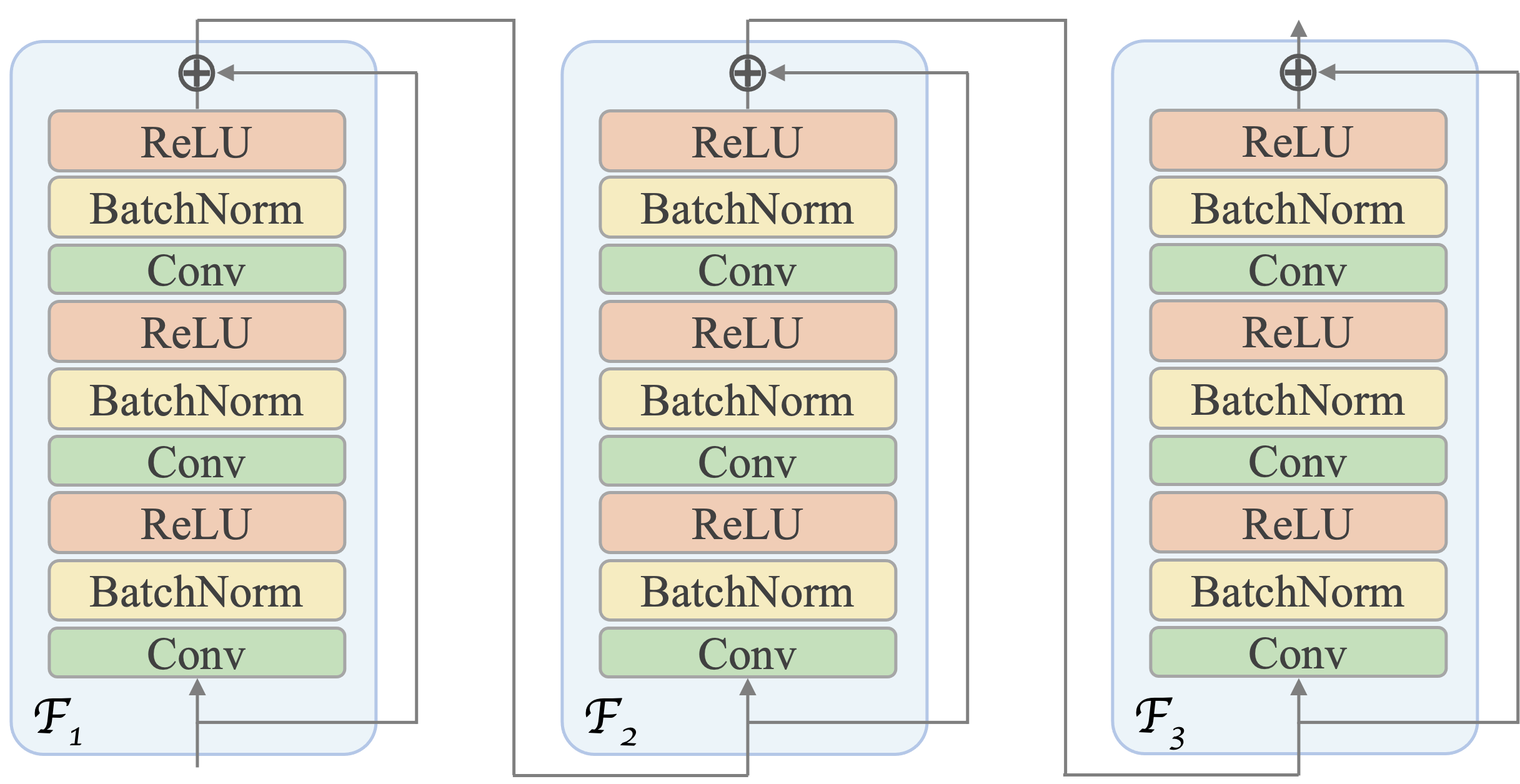}
\end{center}
\caption{\textbf{Illustration of $\mathcal{F}_i$ blocks (with bottleneck) after ReLU layers are moved inside the blocks.}}
\label{fig:Relu_in}
\end{figure}

To deal with this special case, we provide a simple ReLU-in strategy that enables reversibility without affecting module parameters. As shown in Figure~\ref{fig:Relu_in}, we simply move this ReLU layer inside the $\mathcal{F}_i$  block before rewiring. Because $\mathcal{F}_i$ block can be any function, it can include one more ReLU layer, which doesn't have any parameters. After this, we can follow the rewiring process in Section 3.2 in the paper to convert the modules to reversible modules.

Actually, the Slowfast networks we used in the experiments contain such Resnet modules. We followed the above ReLU-in strategy to build the Re$^2$ Slowfast backbones. We showed  that the obtained Re$^2$ Slowfast can reach the performance of the original Slowfast networks in Table~2 in the paper.

\clearpage
{\small
\bibliographystyle{ieee_fullname}
\bibliography{egbib}
}

\end{document}